\pdfoutput=1

\documentclass[11pt]{article}

\usepackage{acl}

\usepackage{times}
\usepackage{latexsym}
\usepackage{todonotes}
\usepackage{amssymb}
\usepackage{pifont}
\newcommand{\cmark}{\ding{51}}%
\newcommand{\xmark}{\ding{55}}%
\usepackage[T1]{fontenc}

\usepackage[utf8]{inputenc}

\usepackage{microtype}

\usepackage{inconsolata}

\usepackage{graphicx}
\usepackage{booktabs}
\usepackage{multirow}
\usepackage{xspace}
\usepackage[normalem]{ulem}
\useunder{\uline}{\ul}{}
\newcommand{\benchmark}{\textsc{CouldAsk}\xspace}
\newcommand{\data}{BanditQA\xspace}

\title{\textit{I Could've Asked That}: Reformulating Unanswerable Questions}

\author{Wenting Zhao
\quad Ge Gao
\quad Claire Cardie \quad Alexander M. Rush\\
Cornell University \\ 
\texttt{\{wz346,gg462,ctc9,arush\}@cornell.edu}}

\begin{document}
\maketitle
\begin{abstract}
    When seeking information from unfamiliar documents, users frequently pose questions that cannot be answered by the documents. While existing large language models (LLMs) identify these unanswerable questions, they do not assist users in reformulating their questions, thereby reducing their overall utility. We curate \benchmark, an evaluation benchmark composed of existing and new datasets for document-grounded question answering, specifically designed to study reformulating unanswerable questions. We evaluate state-of-the-art open-source and proprietary LLMs on \benchmark. The results demonstrate the limited capabilities of these models in reformulating questions. Specifically, GPT-4 and Llama2-7B successfully reformulate questions only 26\% and 12\% of the time, respectively. Error analysis shows that 62\% of the unsuccessful reformulations stem from the models merely rephrasing the questions or even generating identical questions. We publicly release the benchmark\footnote{\url{https://huggingface.co/datasets/wentingzhao/couldask}} and the code to reproduce the experiments\footnote{\url{https://github.com/wenting-zhao/couldask}}.
\end{abstract}

\section{Introduction}
Applying large language models (LLMs) to perform question answering (QA) over documents, such as legal and medical texts, has become increasingly popular~\cite{agrawal-etal-2022-large,guha2023legalbench}.
However, users' limited knowledge of these documents often results in the formulation of unanswerable questions, whose assumptions either conflict with or cannot be verified with the information available in the documents.
We will refer to these assumptions as \emph{presupposition errors}.\footnote{\citet{kim-etal-2023-qa} refer to these assumptions as questionable assumptions.}
\citet{gao-etal-2023-continually} and \citet{yu-etal-2023-crepe} found that around 30\% of information-seeking questions written by users include presupposition errors.
\begin{figure}[t]
  \begin{center}
    \includegraphics[width=0.48\textwidth]{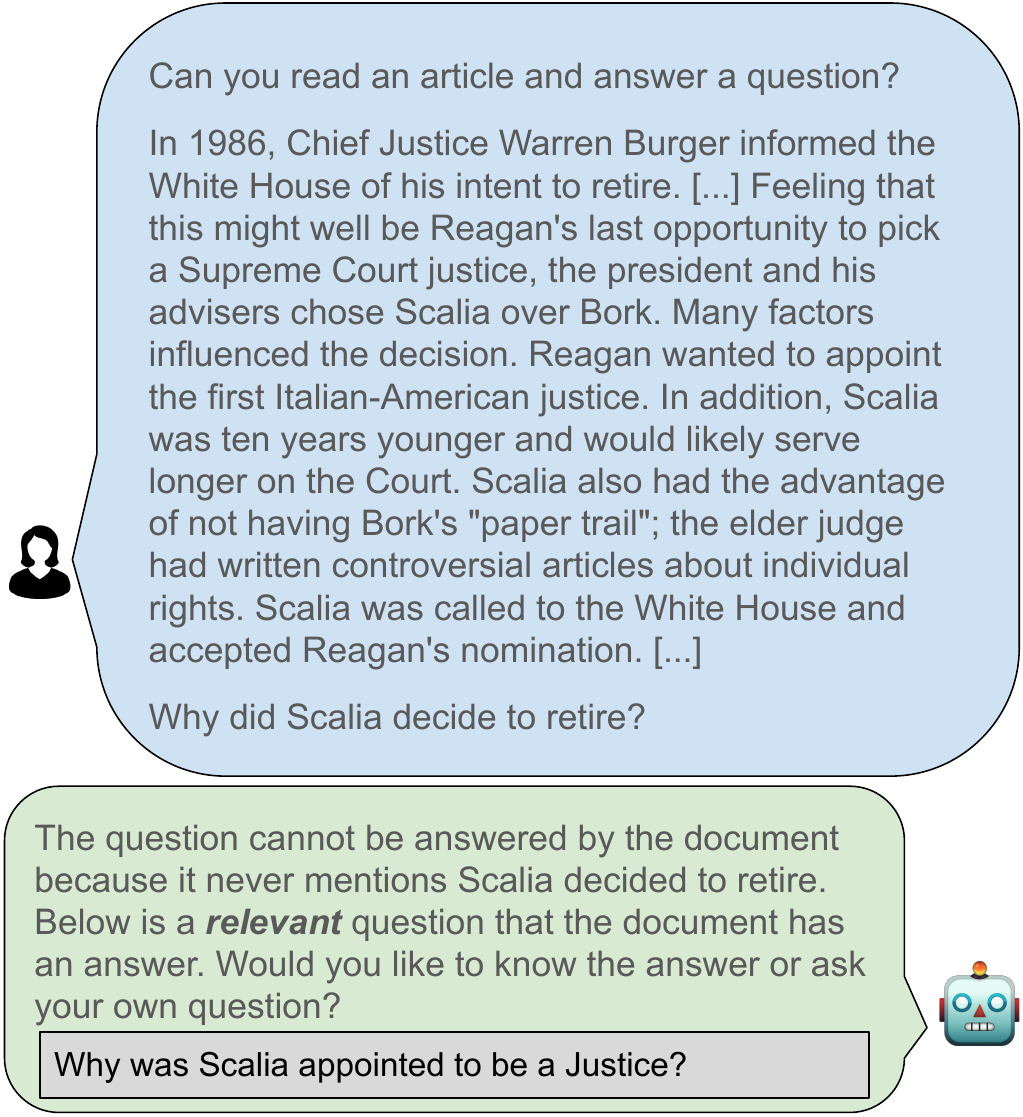}
  \end{center}
  \caption{An example of an LLM suggesting an alternative relevant question the user could have asked whose answers can be found in the document, besides only informing users with the presupposition errors.}
  \label{fig:unanswerable}
\end{figure}
Research in the field has primarily focused on the detection of unanswerable questions \citep{rajpurkar2018know,tran2023agent,hu-etal-2023-wont} and providing explanations for why such questions cannot be answered \citep{yu-etal-2023-crepe,kim-etal-2023-qa}. However, this goal is insufficient for fostering an effective interaction between users and LLMs. Identifying unanswerable questions only serves as a starting point in question reformulation; without additional guidance or feedback on how to rephrase the question, users, especially those unfamiliar with the content, might find themselves caught in a repetitive cycle of formulating questions.
In a large-scale industrial experiment, \citet{faustini-etal-2023-answering} have shown that the practice of rewriting unanswerable questions users ask virtual assistants significantly enhances the experience for millions of users.
\begin{table*}[t]
\centering
\small
\begin{tabular}{p{0.1\textwidth} p{0.07\textwidth} c p{0.32\textwidth} p{0.35\textwidth}}
\toprule
Issue & Strategy & \% & Original question & Revised question \\ \midrule
Contradictory &Correct& 17 &
  What other foods are adulterated with turmeric? &
  What other substances are commonly adulterated in turmeric? \\
                    &&& Which major cities are located at great lakes? & Which U.S. states have jurisdictions that extend into the Great Lakes?              \\  \midrule
Unverifiable & Generalize& 20 &
  When was the Twenty-third Amendment to the United States Constitution passed into law? &
  Has the Twenty-third Amendment to the United States Constitution been proposed? \\
                    &&& How many children does Jennifer Lawrence have? & Is Jennifer Lawrence expecting her first child?                                     \\  \midrule
Unverifiable & Nearest Match & 44 &
  How many arms does Krishna have? &
  How is Krishna typically depicted in terms of arms? \\
                    &&    & How many schools are in CPS?                   & What kind of schools are included in CPS?                                           \\  \midrule
 Unverifiable& Specify &
  \multirow{2}{*}{19} &
  While working at Edison, what inventions did Nikola Tesla make? &
  What specific aspects of electrical engineering did Nikola Tesla work on and further develop while working at Edison? \\
                    &&    & How many children did Toni Morrison have?      & How many children did Toni Morrison have with Harold Morrison before their divorce? \\ \bottomrule
\end{tabular}
\caption{Strategies from a human user reformulating questions on 100 examples.}
\vspace{-10pt}
\label{tab:human-revision}
\end{table*}

This work aims to improve the utility of QA systems by introducing a new task that requires both detecting unanswerable questions and generating questions closely related to the initial query and grounded in the document.
Opting to generate a relevant question rather than a summary of related information emphasizes a user-centered approach. While producing a summary is more document-focused, formulating a relevant question targets understanding and predicting the user's intent, aligning the interaction more closely with the user's specific needs and queries.
We provide an example of how to suggest such questions in Figure~\ref{fig:unanswerable}. Although generating relevant and grounded questions conditioned on initial queries offers greater utility to users, it remains a difficult task even for the best models in a few-shot setting. 



We first characterize human-reformulated questions and describe several different strategies for updating questions to remove presupposition errors.
Motivated by these strategies, we curate \benchmark, an evaluation benchmark for document-grounded QA that consists of a combination of existing and new datasets to study question reformulation in the presence of presupposition errors.
We evaluate several prompting methods such as few-shot prompting and chain-of-thought prompting, employing state-of-the-art open-source and proprietary models. The results illustrate the limitations of existing models and prompting techniques in accurately detecting unanswerable questions and reformulating questions: the F1 scores for identifying unanswerable questions range from 41.16\% to 67.82\%, and success rates for reformulating questions range from 7.13\% to 26.21\%, depending on base models.
Analysis shows that most of the unsuccessful reformulation come from rephrasing or repeating the original questions and that LLMs are worse at reformulating questions necessitating global edits compared to those solely requiring local edits.

\section{Related Work}
Several datasets have been proposed to study unanswerable questions. \citet{rajpurkar2018know} curate the first document-grounded QA dataset that features unanswerable questions.
More recently, \citet{yu-etal-2023-crepe} and \citet{kim-etal-2023-qa} have collected questions with presupposition errors from Google user queries and Reddit posts, respectively.

How to identify unanswerable questions, especially with off-the-shelf LLMs, has remained understudied.
\citet{kim-etal-2021-linguist} proposed to first extract presuppositions from a question and then perform natural language inference (NLI) to check for presupposition violations. However, this pipeline requires supervision.
In practice, supervision is often not available; \citet{kim-etal-2023-qa} thus explore prompting large language models in the chain-of-thought style to identify unanswerable questions.
However, the results remain unsatisfying; using GPT-3 only yields detection accuracy that is only slightly better than random guesses.

\citet{faustini-etal-2023-answering} investigate unanswerable questions and their reformulations in the domain of spoken QA, focusing on issues stemming from disfluencies, grammatical errors, and awkward phrasing. We, however, study unanswerable questions arising from presupposition errors, which require a more profound semantic comprehension of both contexts and questions.

Unanswerable questions are closely related to ambiguous questions \cite{min-etal-2020-ambigqa}. While there has been extensive research into reformulating ambiguous questions \cite{rao-daume-iii-2018-learning, white-etal-2021-open, pyatkin-etal-2023-clarifydelphi,majumder-etal-2021-ask}, the problem of reformulating unanswerable questions receives little attention. Strategies for rephrasing ambiguous questions often involve making questions more specific by mentioning precise entities or events \cite{min-etal-2020-ambigqa}. In contrast, as we will show in Section~\ref{sec:task}, reformulating unanswerable questions necessitates a wide range of strategies.

Finally, we discuss the connection between document-grounded QA and open-domain QA \cite{kim-etal-2023-qa}. Document-grounded QA is essentially open-domain QA with the correct document retrieved. Reformulating questions based on the identified document is a separate skill from retrieving the document. Therefore, question reformulation is an interesting task in itself.

\begin{table*}[t]
\centering
\small
\begin{tabular}{lrcccccc}
\toprule
                  & Total & Unans\% & Document Length & Question Length & \# Question Entities & Source & Domain\\ \midrule
SQuADv2   & 1000 & 50.70 &$143.83 \pm \phantom{0}59.97$&$11.21 \pm 3.54$ & $2.29 \pm 0.94$&Human&Wikipedia\\
QA$^2$ &506&48.81&$818.28\pm684.61$&\phantom{0}$7.74\pm1.39$&$2.05\pm0.57$&Human&Mostly Wikipedia \\
\data & 2070 & 35.56&$261.09\pm145.19$& $\phantom{0}8.37 \pm 2.43$  & $1.70 \pm 0.72$ &Human&Wikipedia\\
BBC     & 278 & 21.22 &$477.75\pm289.41$&$16.37\pm3.49$&$3.31\pm1.04$&GPT-4&News\\
Reddit & 313 & 36.10 &$477.43\pm307.18$&$14.27\pm2.75$&$2.93\pm0.78$&GPT-4&Social Media\\
Yelp & 165 & 30.91 &$387.70\pm147.83$&$15.27\pm3.04$&$3.04\pm0.76$&GPT-4&Review\\ \bottomrule
\end{tabular}
\caption{An overview of \benchmark. Unans\% is the percentage of unanswerable questions.}
\label{tab:datasets}
\end{table*}
\vspace{-5pt}
\section{Task: Question Reformulation}
\label{sec:task}
To assist users with question reformulation when reading unfamiliar documents, we define the following task. Given a document and a user question, the system must determine if the question is unanswerable. Upon identifying the unanswerable question, it must reformulate the question such that the new question is answerable by the document while remaining relevant to the original question.


As this task is challenging to formally define, we begin with a qualitative study over a set of example reformulations by a human user, shown in Table~\ref{tab:human-revision}.
Different strategies are applied for different presupposition errors in the reformulation process.
For handling presuppositions that are contradictory to the documents, the human user corrects the presuppositions.
When it comes to presuppositions that are unverifiable given the documents, we observe three strategies. The first strategy takes a step back by asking about a less specific event than the event in the original question. The second strategy seeks a ``nearest match'' question that the document can answer due to a flaw in the original. The third strategy refines the original question by asking about something more specific that can be verified by the document. While these strategies are not exhaustive, they demonstrate the challenging nature of the problem and the necessity for establishing sources of ground truth in the document.  


\section{The \benchmark Benchmark}
Motivated by the need to reformulate questions both to rely on verified information and to avoid contradictions, we develop an evaluation benchmark. We consider two important challenges in constructing benchmarks for this task. \textbf{(1)} The benchmark should cover a wide range of domains to cover different types of presuppositions. (Existing QA datasets that study unanswerable questions mostly rely on Wikipedia articles~\cite{rajpurkar2018know,gao-etal-2023-continually,kim-etal-2023-qa}.) \textbf{(2)} The evaluation method should be capable of fairly assessing equally good reformulated questions, considering the subjective nature of question reformulation.

\subsection{Datasets}
Following the desiderata, we select three existing datasets -- SQuADv2, \data~\cite{gao-etal-2023-continually}, and QA$^2$~\cite{kim-etal-2023-qa} -- and to cover a broader range of domains, we create three new datasets in the domains of news, review, and Reddit, where the questions are generated by models and verified by crowdworkers. We summarize statistics for all datasets in Table~\ref{tab:datasets}.

\begin{table*}[ht!]
\centering
\small
\begin{tabular}{p{0.43\textwidth} p{0.37\textwidth} c c}
\toprule
Example & Revised question & Ans & Rel\\ \midrule
\multirow{4}{0.43\textwidth}{\textbf{Question}: When did Chick-fil-A open their first restaurant in Pennsylvania?\\\textbf{Document}: [...] he registered the name Chick-fil-A, Inc. From 1964 to 1967, the sandwich was licensed to over fifty eateries, including Waffle House and the concession stands of the new Houston Astrodome. The Chick-Fil-A sandwich was withdrawn from sale at other restaurants when the first standalone location opened in 1967, in the food court of the Greenbriar Mall, in a suburb of Atlanta. Since 1997, the Atlanta-based company has been the title sponsor of the Peach Bowl, an annual college football bowl game played in Atlanta on New Year's Eve.} & $\bullet$ Did the article mention the first Chick-fil-A restaurant in Pennsylvania? & \cmark & \cmark \\
& $\bullet$ Which Chick-fil-A mentioned in the article is the closest to Pennsylvania? & \cmark     & \cmark\\
& $\bullet$When did Chick-fil-A open their first freestanding location? & \cmark     & \xmark   \\
& $\bullet$ How many eateries of Chick-fil-A are licensed from 1964 to 1967? & \cmark     & \xmark   \\
& $\bullet$ Where is the first Chick-fil-A restaurant in Pennsylvania?         & \xmark     & \cmark\\
& $\bullet$ When did Chick-fil-A open its first restaurant in a northern state, specifically in Pennsylvania?         & \xmark     & \cmark   \\
& $\bullet$ When were chicken sandwiches invented at Chick-fil-A?         & \xmark     & \xmark   \\ \bottomrule           
\end{tabular}
\caption{Different aspects of question reformulation. Ans indicates whether the reformulated question can be answered by the document, and Rel indicates whether the reformulated question is relevant to the original question.}
\vspace{-10pt}
\label{tab:metrics}
\end{table*}
\paragraph{BBC, Reddit, and Yelp.} We synthetically generate questions with a question generation model. We use the documents from BBC news articles\footnote{https://huggingface.co/datasets/SetFit/bbc-news}, Reddit pages\footnote{https://huggingface.co/datasets/reddit\_tifu}, and Yelp reviews\footnote{https://huggingface.co/datasets/yelp\_review\_full}, respectively. To not artificially craft unanswerable questions, we instruct the question generation model to produce questions normally and later identify the unanswerable ones. To produce a dataset that is challenging for LLMs, we additionally leverage a question checking model\footnote{GPT-4 is used for both question generation model and question checking model.}. We search for questions that confuse the question checking model. Specifically, we sample the question checking model five times to produce an answer for each of the questions. We gather questions where the question checking model flags the questions are unanswerable any of those five times. We then ask three crowdworkers from Amazon Mechanical Turk (MTurk) to independently verify whether the question is answerable or not\footnote{In particular, to annotate whether a question is answerable, we instruct the crowdworkers to select a span from the document to answer the question. If they cannot identify a span, the question is deemed unanswerable. To reduce noise, we take the majority vote from three crowdworkers. To further ensure a low noise level in annotations, we remove the examples where answer spans annotated by different crowdworkers are disjoint from each other.}. The question generation model produces 9500, 9964, and 10000 QA pairs for the BBC, Reddit, and Yelp datasets, respectively. We keep 278, 313, and 165 examples that have confused the question checking model. Finally, the crowdworkers identify 21.22\%, 36.10\%, and 30.91\% of the questions are truly unanswerable. We thus construct examples that are both challenging to LLMs and high-quality.

\paragraph{SQuADv2, QA$^2$, and \data.} We adapt these datasets to be used for document-grounded QA. Questions in SQuADv2 are mechanistically formulated with the intention of being unanswerable, whereas the other two datasets feature naturally occurring unanswerable questions. QA$^2$ is composed of natural Google search queries, and \data includes questions formulated by users during interactions with LLMs. More information on the modifications we make to these datasets can be found in Appendix~\ref{sec:more-dataset-detail}.

\begin{table*}[t]
\centering
\small
\begin{tabular}{llccccccc}
\toprule
                         &                 & SQuADv2 & QA$^2$ & \data  & BBC  & Reddit & Yelp & Average \\ \midrule
\multirow{4}{*}{GPT-3.5} & ZS & 31.86 & 37.61 & 50.58 & \phantom{0}9.09 & 24.46 & 19.35 & 28.83 \\
 & ZS CoT & 26.49 & 39.52 & 44.74 & 14.08 & 19.40 & 13.33 & 26.26 \\
 & FS & 35.75 & \textit{43.26} & 67.34 & \phantom{0}0.00 & 19.12 & 16.67 & 30.35 \\
 & FS CoT & \textit{51.08} & 37.90 & \textit{70.33} & \textit{16.22} & \textit{37.65} & \textit{33.77} & \textit{41.16} \\\midrule
\multirow{4}{*}{GPT-4} & ZS & \textit{\textbf{85.47}} & 67.40 & 71.24 & 51.93 & 61.82 & 58.74 & 66.10 \\
 & ZS CoT & 84.97 & \textit{68.55} & 67.33 & 52.02 & 55.51 & \textit{\textbf{61.65}} & 65.01 \\
 & FS & 77.34 & 64.17 & 80.48 & 52.07 & 65.95 & 60.74 & 66.79 \\
 & FS CoT & 76.11 & 64.09 & \textit{\textbf{80.85}} & \textit{\textbf{53.93}} & \textit{\textbf{71.48}} & 60.43 & \textit{\textbf{67.82}} \\\midrule
\multirow{4}{*}{Llama2} & ZS & 21.29 & 35.26 & 44.51 & 24.44 & 15.49 & 24.32 & 27.55 \\
 & ZS CoT & 27.76 & 37.39 & 48.22 & 19.61 & 19.23 & 27.16 & 29.89 \\
 & FS & \textit{55.36} & \textit{54.24} & \textit{66.77} & \textit{27.91} & \textit{42.11} & \textit{45.83} & \textit{48.70} \\
 & FS CoT & 35.44 & 35.65 & 58.53 & 25.45 & 35.87 & 34.09 & 37.51 \\\midrule
\multirow{4}{*}{Mistral} & ZS & 38.30 & 30.43 & 46.61 & 30.23 & 26.14 & 17.39 & 31.52 \\
 & ZS CoT & 31.87 & 26.5 & 47.39 & 29.27 & 21.48 & 21.92 & 29.74 \\
 & FS & 41.98 & 34.13 & 47.45 & \textit{38.83} & 42.74 & \textit{40.38} & 40.92 \\
 & FS CoT & \textit{46.65} & \textit{44.32} & \textit{58.23} & 38.10 & \textit{48.28} & 32.97 & \textit{44.76} \\\midrule
\multirow{4}{*}{Zephyr} & ZS & 63.73 & 55.37 & 67.61 & \textit{45.31} & 38.04 & 32.50 & 50.43 \\
 & ZS CoT & 65.42 & 50.36 & 68.27 & 39.72 & 34.07 & 39.53 & 49.56 \\
 & FS & \textit{67.29} & 63.06 & 67.67 & 42.86 & \textit{50.00} & 30.23 & 53.52 \\
 & FS CoT & 64.25 & \textit{\textbf{71.10}} & \textit{68.61} & 42.03 & 49.61 & \textit{43.40} & \textit{56.50}    \\ \bottomrule
\end{tabular}
\caption{Comparing F1 scores for unanswerable question detection with different prompting methods using both proprietary and open-source models on \benchmark. The best method for each base model is italicized, and the best method across all base models is bolded.}
\label{tab:detection}
\vspace{-10pt}
\end{table*}
\begin{table*}[t]
\centering
\small
\begin{tabular}{llccccccc}
\toprule
 &  &SQuADv2 &QA$^2$ & \data & BBC & Reddit & Yelp & Average \\ \midrule
\multirow{4}{*}{GPT-3.5} & ZS & \phantom{0}3.55 & \phantom{0}6.88 & \phantom{0}5.71 & \phantom{0}1.69 & \phantom{0}\textit{4.42} & \phantom{0}\textit{9.80} & \phantom{0}5.34 \\
 & ZS CoT & \phantom{0}3.55 & \phantom{0}3.64 & \phantom{0}4.35 & \phantom{0}1.69 & \phantom{0}2.65 & \phantom{0}3.92 & \phantom{0}3.30 \\
 & FS & \phantom{0}3.35 & \textit{\phantom{0}8.91} & \phantom{0}6.25 & \phantom{0}0.00 & \phantom{0}0.88 & \phantom{0}5.88 & \phantom{0}4.21 \\
 & FS CoT & \textit{\phantom{0}5.52} & \phantom{0}8.50 & \textit{\phantom{0}7.20} & \textit{10.17} & \phantom{0}3.54 & \phantom{0}7.84 & \textit{\phantom{0}7.13} \\\midrule
\multirow{4}{*}{GPT-4} & ZS & \textit{\textbf{17.16}} & \textit{\textbf{23.89}} & \textit{\textbf{11.68}} & \textit{\textbf{52.54}} & \phantom{0}8.85 & \textit{\textbf{43.14}} & \textit{\textbf{26.21}} \\
 & ZS CoT & 15.78 & 18.22 & \phantom{0}8.29 & 35.59 & \textit{\textbf{14.16}} & 35.29 & 21.22 \\
 & FS & 10.45 & 17.00 & 10.19 & 42.37 & \phantom{0}8.85 & 31.37 & 20.04 \\
 & FS CoT & \phantom{0}9.86 & 17.81 & 10.60 & 44.07 & \phantom{0}7.96 & 17.65 & 17.99 \\\midrule
\multirow{4}{*}{Llama2} & ZS & \phantom{0}2.56 & \phantom{0}6.88 & \phantom{0}5.30 & 13.56 & \phantom{0}4.42 & \textit{11.76} & \phantom{0}7.42 \\
 & ZS CoT & \phantom{0}3.75 & \phantom{0}3.24 & \phantom{0}2.99 & 10.17 & \phantom{0}1.77 & \phantom{0}9.80 & \phantom{0}5.29 \\
 & FS & \textit{\phantom{0}7.10} & \textit{14.17} & \textit{\phantom{0}8.15} & \textit{16.95} & \textit{\phantom{0}8.85} & 17.65 & \textit{12.14} \\
 & FS CoT & \phantom{0}3.16 & \phantom{0}4.45 & \phantom{0}5.84 & 13.56 & \phantom{0}3.54 & \phantom{0}5.88 & \phantom{0}6.07 \\\midrule
\multirow{4}{*}{Mistral} & ZS & \phantom{0}1.58 & \phantom{0}2.02 & \phantom{0}4.35 & 11.86 & \phantom{0}1.77 & \phantom{0}1.96 & \phantom{0}3.92 \\
 & ZS CoT & \phantom{0}2.96 & \phantom{0}1.21 & \phantom{0}4.08 & \phantom{0}6.78 & \phantom{0}0.88 & \phantom{0}7.84 & \phantom{0}3.96 \\
 & FS & \phantom{0}3.35 & \textit{\phantom{0}4.05} & \phantom{0}3.67 & \textit{22.03} & \textit{\phantom{0}6.19} & \textit{11.76} & \phantom{0}8.51 \\
 & FS CoT & \textit{\phantom{0}5.92} & \textit{\phantom{0}4.05} & \textit{\phantom{0}6.39} & 16.95 & \phantom{0}5.31 & 13.73 & \textit{\phantom{0}8.72} \\\midrule
\multirow{4}{*}{Zephyr} & ZS & \phantom{0}7.89 & 12.15 & \phantom{0}7.61 & 30.51 & \phantom{0}4.42 & \phantom{0}9.80 & 12.06 \\
 & ZS CoT & \phantom{0}8.09 & \phantom{0}7.69 & \phantom{0}7.34 & 20.34 & \phantom{0}6.19 & \phantom{0}7.84 & \phantom{0}9.58 \\
 & FS & \phantom{0}5.33 & \textit{15.38} & \phantom{0}7.07 & 13.56 & \phantom{0}4.42 & \phantom{0}7.84 & \phantom{0}8.93 \\
 & FS CoT & \textit{\phantom{0}8.88} & 13.77 & \textit{10.19} & \textit{35.59} & \textit{\phantom{0}7.96} & \textit{13.73} & \textit{15.02}      \\ \bottomrule
\end{tabular}
\caption{Question Reformulation. Success rates using different prompting methods with both proprietary and open-source models on \benchmark. The best method for each base model is italicized, and the best method across all base models is bolded.}
\vspace{-10pt}
\label{tab:success-rate}
\end{table*}
\subsection{Evaluating Reformulation}
To improve the utility of QA systems in responding to unanswerable questions, the reformulated questions must be (1) answerable by the documents and (2) relevant to the original questions posed by the users. As illustrated by the examples in Table~\ref{tab:metrics}, a reformulation could be answerable but not relevant, or relevant but not answerable. A successful reformulation must satisfy both conditions.
There are multiple equally good reformulations for each question. For instance, when given the original question, it is hard to determine which of the first two reformulations in Table~\ref{tab:metrics} would be closer to the user's intent. As a result, we opt for a reference-free evaluation approach, where the evaluation does not rely on gold reformulations.

Measuring the relatedness of two questions involves determining how closely their topics, intents, and meanings are aligned. With these goals in mind, we propose three reference-free relevance metrics: edit distance, entity overlap ratio\footnote{We justify our choice of entity overlap ratio as a relevance metric with human evaluation in Section~\ref{sec:analysis}.}, and cosine similarity between the original question and the reformulation to indicate the level of relevance.
We consider the reformulation to be unanswerable and have zero relatedness for the unanswerable questions that are not successfully detected by the system.
To automatically evaluate (1), we train a Llama2-7B model on \benchmark to classify whether the reformulation is answerable or not. The classifier achieved 95\% accuracy on a held-out validation set. We release the classifier on Hugging Face
. For (2), we calculate the Levenshtein edit distance, use GPT-4 to tag entities for computing entity overlap ratios, and apply OpenAI embedding models to produce question embeddings for computing cosine similarities.
Finally, a reformulation that is answerable but irrelevant, or vice versa, is not yet helpful. To consolidate the evaluation into a single unified score, using entity overlap ratios as an example, we assign a score of 1 to a reformulation if it is both answerable and has an entity overlap ratio of more than 50\%; otherwise, we assign a score of 0. We refer to this binary score as \emph{success rate}.

\section{Experimental Setup}
\label{sec:experiments}
\paragraph{Models.} We test a range of instruction-finetuned LLMs. For proprietary models, we consider GPT-3.5 (gpt-3.5-turbo-0125) and GPT-4 (gpt-4-0613). For open-source models, we consider a list of 7-billion-parameter models: Llama2\footnote{huggingface.co/meta-llama/Llama-2-7b-chat-hf} \cite{touvron2023llama}, Mistral\footnote{huggingface.co/mistralai/Mistral-7B-Instruct-v0.2} \cite{jiang2023mistral}, and Zephyr\footnote{huggingface.co/HuggingFaceH4/zephyr-7b-beta} \cite{tunstall2023zephyr}.

\paragraph{Comparisons.} We consider several prompting approaches: zero-shot (ZS) and few-shot (FS) prompting and ZS and FS chain-of-thought (CoT) prompting. We first prompt LLMs to determine whether the input question cannot be answered by the provided document. For ZS and FS prompting, we use the prompt provided by SurgeAI\footnote{SurgeAI prompt: ``Answer the following from the above passage alone, and if you can't determine the answer based on the passage, say that you don’t know the answer.''}, which explicitly instructs the model to not produce an answer if the answer cannot be determined from the document alone. For ZS and FS CoT prompting, we expand the aforementioned prompt by asking the model to think step by step to come up with a reason to explain and support its decision.
Only questions determined to be unanswerable proceed to the question reformulation stage.
In this stage, all methods are provided with their previous turns where the models determine the questions are unanswerable.
ZS and FS prompting instruct the model to make minimum edits to the original question to make it answerable.
ZS and FS CoT prompting instruct the model to think step by step to reason about the minimum edits they can make.
\begin{table*}[t]
\centering
\small
\begin{tabular}{p{0.15\textwidth} c p{0.32\textwidth} p{0.36\textwidth}}
\toprule
Error Category & \% & Original question & Revised question \\ \midrule
\multirow{2}{0.15\textwidth}{Simply rephrasing / producing the same questions} & 62 &
  When was he born? &
  When was Jay Chou born and raised in Taipei, Taiwan? \\
 &
   &
  Which team won the most NFL Europe titles? &
  Who was the team that won the most NFL Europe titles? \\ \midrule
\multirow{2}{0.15\textwidth}{Producing questions by copying documents} & 14 &
  When did the Cubs play their first home game at Wrigley Field? &
  When did the Cubs move to Wrigley Field? \\
 &
   &
  what are the characteristics of the Classical Roman architecture? &
  What are some of the characteristics of Classical Roman architecture that were adopted by Renaissance architects? \\ \midrule
\multirow{2}{0.15\textwidth}{Producing irrelevant questions} & 17 &
  What is Snow Patrol's most famous song? &
  Who are some famous musicians that have expressed admiration for Snow Patrol? \\
 &
   &
  which campus is the biggest in University of California? &
  Which college in the University of California system is not controlled by the Regents or President? \\ \midrule
Other &
  11 &
  How many arms does Krishna have? &
  I apologize for the confusion. The passage does not mention how many arms Krishna has. \\ \bottomrule
\end{tabular}
\caption{Error analysis on 100 revised questions generated by different models.}
\vspace{-10pt}
\label{tab:error-analysis}
\end{table*}

\section{Results}
\paragraph{Detecting Unanswerable Questions} Being able to detect unanswerable questions is a necessary precondition for successful question reformulation. Table \ref{tab:detection} presents the F1 scores for unanswerable question detection using different prompting approaches with each base model.
Performance is often better for existing datasets than new ones, which indicates our approach's effectiveness in generating more challenging questions.
Among all models, GPT-4 performs best at identifying unanswerable questions.
Surprisingly, both Mistral and Zephyr are more accurate at detecting unanswerable questions than GPT-3.5.
Among all prompting techniques, FS CoT consistently improves upon ZS, with a larger degree of improvement observed in smaller models compared to larger ones.

\paragraph{Reformulating Questions} Table \ref{tab:success-rate} presents success rates for question reformulation vary dramatically from domain to domain. News (BBC) appears to be the least challenging domain, with success rates ranging from 10.17 to 52.54 depending on base models. Reddit is a challenging domain, with success rates ranging from 4.42 to 14.16. The results on Wikipedia are mixed. While SQuADv2, QA$^2$, and \data are in the Wikipedia domain, LLMs achieve the lowest success rates on \data. We hypothesize that user queries written during interaction require deeper revision.

Among all base models, GPT-4 achieves the highest average success rate (26.21), while GPT-3.5 has the lowest average success rate (7.13). Among all open-source models, Zephyr has the best performance. When it comes to prompting methods, there is not a clear winner. Different LLMs can be improved with different prompting approaches. GPT-3.5, Mistral, and Zephyr benefit the most from FS CoT prompting, GPT-4 from ZS, Llama2 from FS. We include individual metrics for question reformulation in Appendix~\ref{sec:more-reformulation}.
\section{Analysis}
\label{sec:analysis}


\paragraph{Qualitative analysis}
We randomly sample and analyze 20 reformulated questions from each base model (a total of 100 questions) that cannot be answered by the corresponding documents. We summarize the results in Table~\ref{tab:error-analysis}. We identify three major types of errors. The most frequent type is that the models simply rephrase or generate the same questions. Most of the errors in this type are contributed by open-source models such as Llama2.

Another type of error that occurs 14\% of the time is that the models generate a question by copying a document span that looks similar to the original question. For example, given the original question about the characteristics of the Classical Roman architecture, the model identifies a span that is similar to the original question and generates a question based on this span. However, the model does not plan thoroughly about whether the new question can be answered or not.

For 17\% of the time, the models propose questions that are irrelevant to the original questions. For example, in the Snow Patrol question, the subject of the reformulation has entirely changed from asking about the most famous song to famous musicians. Finally, we group the remaining eleven errors in the Other type, where there usually is not even a question produced.

\begin{figure}[t!]
    \centering
    \framebox{
    \parbox{0.45\textwidth}{
    \small
    \textbf{Document:}\\In December 2007, Sport England published a survey which revealed that residents of St Albans were the 10th most active in England in sports and other fitness activities. 30.8\% of the population participate at least 3 times a week for 90 minutes.\\
    \textbf{Question:}\\What is St Albans Sport?\noindent \smallskip \newline
    \textbf{ZS CoT for unanswerable question detection:}\\The article did not provide information on what St Albans Sport is, so I don't know.\\
    \textbf{ZS CoT for question reformulation:}\\The article discusses the activity level of the residents of St Albans in sports and fitness but does not detail a specific sport. Instead of asking "What is St Albans Sport?" we could ask, "How often do residents of St Albans participate in sports and fitness activities?" to get an answer from the given passage.\noindent \smallskip \newline
    \textbf{ZS for question reformulation:}\\What is the level of participation in sports and other fitness activities in St Albans?}
    }
    \caption{Example outputs produced by GPT-4. Via prompting, the model detects unanswerable questions then reformulates the questions with a second prompt.}
    \label{fig:gpt4-example}
\end{figure}

\paragraph{Limited benefit from FS CoT on Question Reformulation.} We explore why FS CoT improves the detection of unanswerable questions but not question reformulation by conducting a qualitative analysis of GPT-4 outputs on a \data example, as illustrated in Figure~\ref{fig:gpt4-example}. It is relatively straightforward to determine why a question is unanswerable --- either the document does not provide the necessary information or there is a presupposition that conflicts with the document. However, question reformulation demands compositional reasoning. The model needs to first decide on a reformulation strategy and then plan the specific steps to achieve the reformulation. The reformulated question generated by GPT-4 via FS CoT prompting is closer to being answerable. However, the model misses a subtle detail --- the document only mentions the exercise habits of 30.8\% of residents, not the general population. Therefore, FS CoT alone, without further methodological innovation, does not fully address the challenge of question reformulation.

\begin{figure*}[t]
    \centering
    \includegraphics[width=\textwidth]{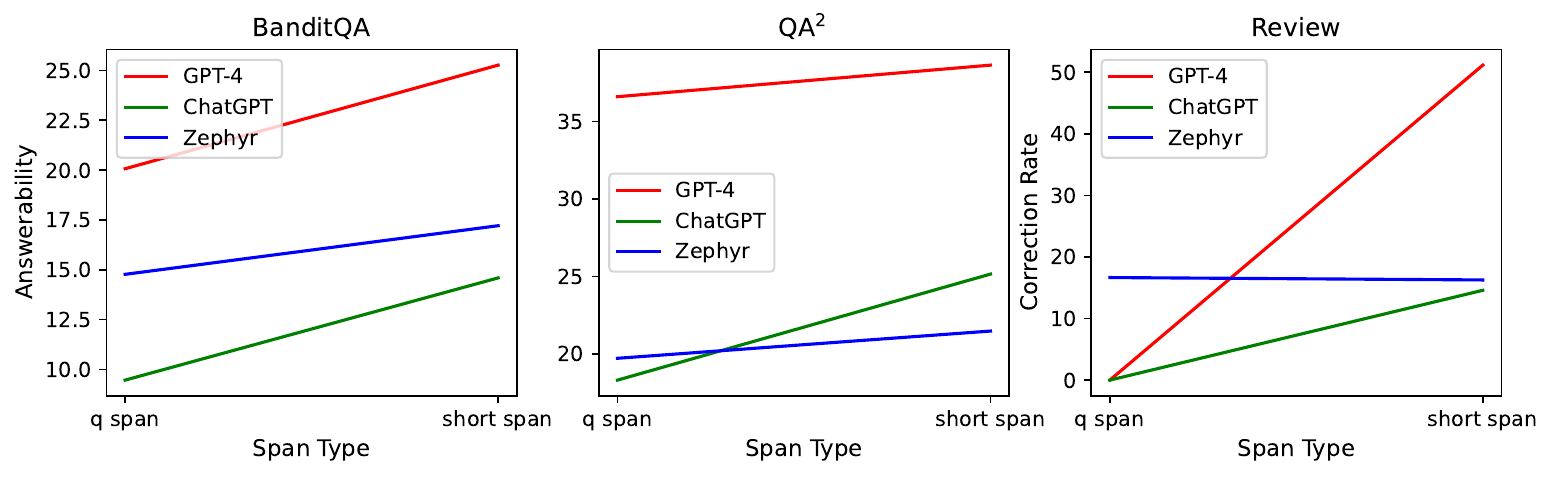}
    \caption{The relation between the percentage of reformulations answerable by the documents and the type of minimum spans in the original questions. q span are the examples where the minimum span is the full question, while short span includes all other instances.}
    \label{fig:compositionality}
\end{figure*}

\begin{table}[t]
    \centering
    \small
    \begin{tabular}{l p{0.3\textwidth}}
    \toprule
        \multicolumn{2}{c}{q span} \\ \midrule
        Original Q & \textbf{What medicine is made from Coca?} \\
        Reformulated Q & What substance is derived from the Coca leaf? \\ \toprule
        \multicolumn{2}{c}{short span} \\ \midrule
        Original Q & When was \textbf{USB-C} developed? \\
        Reformulated Q & When was USB developed? \\ \bottomrule
    \end{tabular}
    \caption{Examples of a question with the minimum span being the entire question (q span) and a question where the minimum span is a noun (short span).}
    \label{tab:span}
\end{table}

\paragraph{Compositional Modifications vs. Answerability.}
We hypothesize that it is more difficult to reformulate an unanswerable question to be answerable when it requires compositional modifications, which means making global edits to the question instead of making local edits. 

To quantify edits required, we follow \citet{lee-etal-2020-squad2} and annotate the minimum span of a question to explain why the question cannot be answered by the document. We use GPT-4 to annotate unanswerable questions in \data, QA$^2$, and Yelp.
We divide the questions into two categories. The first category comprises questions whose minimum span is the entire question. We call these \emph{q span}. The second category covers questions that have a shorter minimum span, typically a noun or a noun phrase. We call these \emph{short span}. We calculate how many reformulations are answerable, broken down by the type of minimum span.
We consider the reformulations generated by zero-shot prompting with GPT-4, GPT-3.5, and Zephyr. 

The results of this analysis are in Figure~\ref{fig:compositionality}. Our findings consistently show that fewer reformulations are answerable when the minimum span constitutes the entire question compared to when the minimum span is shorter, with only one exception in zero-shot prompting with Zephyr on Yelp.

We present examples for each span type in Table~\ref{tab:span}. When the minimum span is the entire question, we cannot attribute the presupposition error to a smaller segment of the question. As a result, the modification has to be applied to the full question. For example, for the q-span example in Table~\ref{tab:span}, modifying individual parts of the question cannot correct the presupposition error, and therefore the model needs to make global changes. On the contrary, when the minimum span is short, it only requires local edits. For the short-span question in Table~\ref{tab:span}, replacing USB-C with USB is enough to correct the question. Future efforts should be more devoted to the more challenging cases to make progress on question reformulation.

\begin{table}[t]
    \centering
    \small
    \begin{tabular}{lcc}
    \toprule
       & Edit Distance & Entity Overlap \\ \midrule
       Fleiss' $\kappa$  & 40.96 & 93.99 \\
       \bottomrule
    \end{tabular}
    \caption{The correlation between the tested metrics and human-judged relevance between two questions, evaluated using Fleiss' $\kappa$.}
    \label{tab:correlation}
\end{table}

\paragraph{Sufficiency of entity overlap ratios.} 
Our metric is for relevance is based on a minimum entity
overlaps. To judge the sufficiency of this metric, we 
compare it to human evaluation. As a baseline we consider Levenshtein edit distance, a method to measure the similarity between two questions, as a baseline metric. We have a human annotator evaluate 200 pairs of reformulated questions produced by zero-shot prompting with GPT-4 and GPT-3.5 on \data. The question pairs are randomly shuffled to ensure the annotator remains unaware of the model source of each reformulation. The annotator then selects the more relevant question from each pair, based on alignment with the original question's intent. Subsequently, we identify the question with a higher entity overlap ratio and the question with a lower edit distance as the more relevant ones, respectively. 


We calculate the Fleiss' $\kappa$ score between human evaluations and each of the metrics, with the results summarized in Table~\ref{tab:correlation}. Compared to edit distance, the entity overlap ratio more accurately represents the relevance between the original and reformulated questions (93.99 vs. 40.96). A Fleiss' $\kappa$ score of 93.99 also suggests a near-perfect agreement between the entity overlap ratio and human judgements.  We hypothesize that the specific semantic properties of the questions are more central than the surface form represented by edit distance. 


\section{Conclusion}
Users often ask unanswerable questions when they seek information from unfamiliar documents. 
Existing LLMs identify these questions but do not aid users in reformulating new questions, resulting in ineffective user-LLM interactions.
We introduce \benchmark, a compiled set of grounded-document QA datasets designed to study both unanswerable question detection and question reformulation.

\section*{Limitations}
While our benchmark offers advantages over existing sources, we acknowledge the following limitations.
Questions in BBC, Reddit, and Yelp are generated by GPT-4, and they may not accurately represent questions posed by humans.
Despite best efforts to ensure high-quality annotations, occasional human errors are possible.
Additionally, our benchmark only collects English questions and thus lacks language diversity.
Finally, regarding evaluation, the way we currently measure success rates only focuses on mistakes made on unanswerable questions. If an answerable question is detected to be unanswerable, we do not evaluate question reformulation in such cases.

\section*{Acknowledgments}

This work was supported by NSF CAREER \#2037519 and NSF \#1901030. We also thank Lillian Lee and the anonymous reviewers for their helpful feedback.

\bibliography{anthology,custom}
\bibliographystyle{acl_natbib}

\appendix
\begin{table*}[t]
\small
\centering
\begin{tabular}{lccccccc}
\toprule
                   & SQuADv2 & QA$^2$   & BanditQA & BBC   & Reddit & Yelp  & Average \\ \midrule
Prompting          & 18.74   & 21.86 & 10.60     & 47.46 & 7.96   & 33.33 & 23.33   \\
Explicit Prompting & 11.43   & \phantom{0}8.91  & \phantom{0}4.07     & 18.64 & 7.07   & 17.64 & 11.00   \\
Rule-based         & \phantom{0}4.50    & \phantom{0}2.43  & \phantom{0}1.22     & 11.86 & 0.88   & \phantom{0}3.90  & \phantom{0}4.13   \\
\bottomrule
\end{tabular}
\caption{Comparing the standard zero-shot prompting method to various baselines to hack the proposed metric, success rates.}
\label{tab:hack}
\end{table*}

\begin{table*}[t]
\centering
\small
\begin{tabular}{llccccccc}
\toprule
                                      &                    & SQuADv2 & QA$^2$   & BanditQA & BBC   & Reddit & Yelp  & Average \\ \midrule
\multirow{2}{*}{Answerability}        & Prompting          & 28.01   & 36.84 & 36.84    & 71.19 & 13.27  & 47.06 & 36.58   \\
                                      & Rule-based & 28.01   & 35.22 & 24.86    & 62.71 & 10.61  & 52.94 & 35.73   \\
\multirow{2}{*}{Entity Overlap Ratio} & Prompting          & 38.64   & 24.87 & 29.13    & 45.00 & 44.73  & 48.69 & 38.51   \\
                                      & Rule-based & 21.36   & 13.60 & 13.15    & 23.50 & 15.55  & 21.41 & 18.10  \\ \bottomrule
\end{tabular}
\caption{Analysis of individual metrics used to compute success rates.}
\label{tab:hack-analysis}
\end{table*}
\section{Dataset Details}
\label{sec:more-dataset-detail}
We provide more details about the existing datasets we include in \benchmark.
\paragraph{SQuADv2} is a collection of QA pairs from Wikipedia articles, including questions with presupposition errors. Such questions are written by crowdworkers who are instructed to craft questions that are plausible but have presuppositions that cannot be verified by the associated articles. These annotators either modify valid questions or create new questions based on entities or topics related to the text, ensuring the questions appear relevant and deceivingly valid.

\paragraph{QA$^2$} contains QA pairs from general web articles. QA$^2$ sources questions from Google’s autocompletion API. It calls the API on prefix strings -- when, where, which, how, what, why, who, whose -- to complete the queries. Crowdworkers are recruited to (1) search for relevant documents and (2) check whether these questions contain presupposition errors. QA$^2$ features naturally occurring presupposition errors and diverse document sources. The original QA$^2$ dataset is open-domain QA and only provides annotations for relevant URLs. To obtain the context, we scrape the text from their annotated URL and remove the examples where the contexts are not clear.

\paragraph{\data} investigate whether continual human feedback can improve extractive QA systems; we repurpose the dataset to study naturally occurring presupposition errors. During data collection, users are presented with Wikipedia passages, and they are required to write questions about things they are curious to know. In a later verification step, over 30\% of these questions are identified to have presupposition errors. \data is closest to the setting of this study -- when users are unfamiliar with the documents, they come up with presuppositions that cannot be grounded in the documents.

\section{More Analysis}
\label{sec:more-reformulation}
\paragraph{Hacking success rates.} We explore the potential for our proposed metrics to be manipulated in order to artificially achieve perfect scores. We consider two approaches. The first approach directly prompts LLMs to produce an answerable question while leaving all the entities unchanged. In the second approach, we test a rule-based heuristic that has the following steps: (1) tag entities in the original question via LLM prompting, (2) select the sentence in the document that has the highest entity overlap ratio with the original question, and (3) replace something in the select sentence to create a wh-question, again using LLM prompting. The results, summarized in Table~\ref{tab:hack}, indicate that our success rates remain robust, even when faced with methods explicitly designed to exploit our metric.

We seek to understand where performance degrades in a rule-based heuristic by inspecting individual metrics used to compute success rates. We present the results in Table~\ref{tab:hack-analysis}. While using the rule-based heuristic leads to similar answerability performance, the entity overlap ratio drops significantly. We show further qualitative analysis of the error types in Figure~\ref{fig:hack-qualitative}.

\begin{figure}[t]
    \centering
    \framebox{
    \parbox{0.45\textwidth}{
    \small
    \textbf{Failure case 1: Not all entities present in the same sentence}\\
    \textbf{Original unanswerable question:} What are balance in an open system of particles?\\
    \textbf{Entities:} balance, open system, particles\\
    \textbf{Most-overlapped sentence in the document:} This means that in a closed system of particles, there are no internal forces that are unbalanced.\\
    \textbf{Reformulated Question:} What does this mean about internal forces in a closed system of particles?\\
    \textbf{Note}: The three entities are mentioned in different sentences. Therefore, the most overlapped sentence only has one overlapped entity. Although this question can be answered, the reformulation is not successful due to the low entity overlap ratio. Note that most of our questions have an average number of two to three entities.\noindent \smallskip \newline
    \textbf{Failure case 2: Reformulated questions are still unanswerable}\\
    \textbf{Original unanswerable question:} What job requires no qualifications?\\
    \textbf{Entities:} job, qualifications \\
    \textbf{Most-overlapped sentence in the document:} As in the House of Commons, a number of qualifications apply to being an MSP.\\
    \textbf{Reformulated Question:} What qualifications apply to being an MSP as in the House of Commons?\\
    \textbf{Note}: Even though a part of the sentence was replaced with a what question, this process does not necessarily make the reformulated question become answerable. The article does not mention what specific qualifications, and successful reformulations require a deeper understanding of the document as a whole.
    }
    }
    \caption{Qualitative analysis on the rule-based heuristic approach to hack our proposed metric, success rates.}
    \label{fig:hack-qualitative}
\end{figure}

\paragraph{Individual Metrics in Question Reformulation} We present individual metrics for question reformulation. Table \ref{tab:answerable} presents the answerability (top) and entity overlap ratios (bottom) achieved by each method using each base model. We additionally report cosine similarities in Table~\ref{tab:cosine-similarity} and edit distance in Table~\ref{tab:edit-distance}.

\paragraph{Assessing the Accuracy of Entity Tagging Models through Human Evaluation} To determine the entity overlap ratio, we first utilize GPT-4 to identify entities within both the original and revised questions. To ensure the credibility of the tagged entities, we conduct a human evaluation on a set of 100 questions. The findings of this evaluation are shown in Table~\ref{tab:entity-tagging}. This analysis reveals that GPT-4 exhibits both high precision and recall in the identification of entities, affirming its effectiveness for this task.

\begin{table}[t]
    \centering
    \small
    \begin{tabular}{lcc}
    \toprule
         & Precision & Recall \\ \midrule
        GPT-4 & 99 & 94 \\ \bottomrule
    \end{tabular}
    \caption{Performance of applying GPT-4 to tag entities.}
    \label{tab:entity-tagging}
\end{table}

\begin{table}
    \centering
    \small
    \begin{tabular}{lccc}
    \toprule
         & BBC & Reddit & Yelp \\ \midrule
         Fleiss’ $\kappa$ &65.07 & 59.94 & 58.76\\ \bottomrule
    \end{tabular}
    \caption{Inter-annotator agreement rates among three workers for annotating unanswerable questions.}
    \label{tab:human-agreement}
\end{table}

\begin{table*}[htbp]
\centering
\small
\begin{tabular}{llccccccc}
\toprule
                         &                 & SQuADv2 & QA$^2$ & \data  & BBC  & Reddit & Yelp & Average \\ \midrule
                         &&\multicolumn{6}{c}{Answerability}\\ \midrule
\multirow{4}{*}{GPT-3.5} & ZS & \phantom{0}4.73 & 12.55 & \phantom{0}8.42 & \phantom{0}5.08 & \textit{\phantom{0}4.42} & \phantom{0}9.80 & \phantom{0}7.50 \\
 & ZS CoT & \phantom{0}5.92 & 10.93 & \phantom{0}8.29 & \phantom{0}5.08 & \phantom{0}2.65 & \phantom{0}5.88 & \phantom{0}6.46 \\
 & FS & \phantom{0}7.30 & \textit{18.22} & 14.54 & \phantom{0}0.00 & \phantom{0}0.88 & \phantom{0}5.88 & \phantom{0}7.80 \\
 & FS CoT & \textit{\phantom{0}9.07} & 16.19 & \textit{15.35} & \textit{10.17} & \phantom{0}3.54 & \textit{11.76} & \textit{11.02} \\\midrule
\multirow{4}{*}{GPT-4} & ZS & 26.82 & \textit{\textbf{36.84}} & \textbf{22.96} & \textit{\textbf{72.88}} & \phantom{0}8.85 & \textit{\textbf{58.82}} & \textit{\textbf{37.86}} \\
 & ZS CoT & \textit{\textbf{27.22}} & 34.82 & 19.16 & 66.10 & \textit{\textbf{14.16}} & 52.94 & 35.73 \\
 & FS & 21.10 & 34.01 & \textit{25.41} & 66.10 & \phantom{0}8.85 & 47.06 & 33.76 \\
 & FS CoT & 21.30 & 34.82 & 24.18 & 71.19 & \phantom{0}7.96 & 52.94 & 35.40 \\\midrule
\multirow{4}{*}{Llama2} & ZS & \phantom{0}4.73 & 12.55 & \phantom{0}8.15 & 18.64 & \phantom{0}4.42 & 11.76 & 10.05 \\
 & ZS CoT & \phantom{0}5.13 & \phantom{0}5.67 & \phantom{0}5.30 & 13.56 & \phantom{0}1.77 & \phantom{0}9.80 & \phantom{0}6.87 \\
 & FS & \textit{13.41} & \textit{25.51} & \textit{12.36} & \textit{28.81} & \textit{10.62} & \textit{23.53} & \textit{19.04} \\
 & FS CoT & \phantom{0}5.13 & \phantom{0}8.50 & \textit{\phantom{0}8.42} & 15.25 & \phantom{0}5.31 & \phantom{0}7.84 & \phantom{0}8.41 \\\midrule
\multirow{4}{*}{Mistral} & ZS & \phantom{0}3.55 & \phantom{0}6.48 & \phantom{0}6.39 & 15.25 & \phantom{0}2.65 & \phantom{0}3.92 & \phantom{0}6.37 \\
 & ZS CoT & \phantom{0}5.13 & \phantom{0}2.02 & \phantom{0}7.74 & 11.86 & \phantom{0}2.65 & \phantom{0}7.84 & \phantom{0}6.21 \\
 & FS & \phantom{0}6.90 & \phantom{0}8.50 & \phantom{0}7.47 & \textit{28.81} & \textit{\phantom{0}6.19} & \textit{19.61} & 12.92 \\
 & FS CoT & \textit{10.85} & \textit{10.12} & \textit{\phantom{0}9.92} & 23.73 & \phantom{0}\phantom{0}5.31 & \textit{19.61} & \textit{13.26} \\\midrule
\multirow{4}{*}{Zephyr} & ZS & 15.78 & 23.89 & 12.77 & \textit{42.37} & \phantom{0}4.42 & 11.76 & 18.50 \\
 & ZS CoT & 14.20 & 12.96 & 12.23 & 33.90 & \phantom{0}6.19 & 11.76 & 15.21 \\
 & FS & \textit{20.32} & \textit{34.82} & \textit{16.98} & 38.98 & \phantom{0}4.42 & 15.69 & \textit{21.87} \\
 & FS CoT & 15.38 & 25.91 & 15.49 & 38.98 & \textit{\phantom{0}7.96} & \textit{21.57} & 20.88   \\ \midrule                   
&&\multicolumn{6}{c}{Entity Overlap Ratios}\\ \midrule
\multirow{4}{*}{GPT-3.5} & ZS & 10.44 & \textit{10.56} & 19.82 & \phantom{0}2.37 & 11.28 & \phantom{0}8.50 & 10.50 \\
 & ZS CoT & \phantom{0}7.27 & \phantom{0}5.94 & 11.39 & \phantom{0}1.81 & \phantom{0}7.82 & \phantom{0}3.27 & \phantom{0}6.25 \\
 & FS & \phantom{0}6.75 & 10.32 & 23.38 & \phantom{0}0.00 & \phantom{0}5.68 & \phantom{0}9.15 & \phantom{0}9.21 \\
 & FS CoT & \textit{14.97} & \phantom{0}8.03 & \textit{28.17} & \textit{\phantom{0}7.91} & \textit{16.74} & \textit{15.20} & \textit{15.17} \\\midrule
\multirow{4}{*}{GPT-4} & ZS & \textit{\textbf{38.73}} & \textit{\textbf{24.93}} & 31.26 & \textit{\textbf{49.49}} & 45.13 & \textit{\textbf{48.86}} & \textit{\textbf{39.73}} \\
 & ZS CoT & 29.99 & 20.28 & 23.73 & 34.97 & 32.98 & 38.30 & 30.04 \\
 & FS & 26.13 & 18.42 & 31.66 & 43.98 & 45.84 & 47.32 & 35.56 \\
 & FS CoT & 24.85 & 18.76 & \textit{33.93} & 41.27 & \textit{\textbf{48.27}} & 39.51 & 34.43 \\\midrule
\multirow{4}{*}{Llama2} & ZS & \phantom{0}5.78 & \phantom{0}8.40 & 18.07 & 12.43 & \phantom{0}6.98 & 12.25 & 10.65 \\
 & ZS CoT & \phantom{0}6.53 & \phantom{0}5.16 & 13.45 & \phantom{0}7.37 & \phantom{0}5.84 & 10.62 & \phantom{0}8.16 \\
 & FS & \textit{21.09} & \textit{17.88} & \textit{35.15} & \textit{14.58} & \textit{30.22} & \textit{31.86} & \textit{25.13} \\
 & FS CoT & \phantom{0}8.16 & \phantom{0}7.42 & 17.53 & 12.74 & 10.34 & 12.42 & 11.43 \\\midrule
\multirow{4}{*}{Mistral} & ZS & \phantom{0}6.85 & \phantom{0}4.93 & 10.73 & 10.88 & \phantom{0}7.67 & \phantom{0}4.41 & \phantom{0}7.58 \\
 & ZS CoT & \phantom{0}6.27 & \phantom{0}2.63 & 11.38 & \phantom{0}6.10 & \phantom{0}4.57 & \phantom{0}9.31 & \phantom{0}6.71 \\
 & FS & \phantom{0}8.88 & \phantom{0}5.74 & 12.99 & \textit{19.07} & \textit{26.55} & \textit{26.63} & \textit{16.64} \\
 & FS CoT & \textit{10.46} & \textit{\phantom{0}6.07} & \textit{16.66} & 15.37 & 22.05 & 20.26 & 15.15 \\\midrule
\multirow{4}{*}{Zephyr} & ZS & 21.25 & 14.24 & 32.94 & 26.50 & 17.92 & 16.01 & 21.48 \\
 & ZS CoT & 17.13 & 11.57 & 23.01 & 19.58 & 14.90 & 17.32 & 17.25 \\
 & FS & 12.19 & 14.88 & 28.60 & 12.43 & 18.07 & 12.81 & 16.50 \\
 & FS CoT & \textit{21.37} & \textit{16.94} & \textbf{\textit{35.29}} & \textit{30.42} & \textit{33.26} & \textit{25.98} & \textit{27.21}\\ \bottomrule
\end{tabular}
\caption{Question reformulation quality broken down by answerability (top) and entity overlap ratios (bottom) on \benchmark. The best method for each base model is highlighted with an underscore, and the best method across all base models is bolded.}
\label{tab:answerable}
\end{table*}

\begin{table*}[t]
\centering
\small
\begin{tabular}{llccccccc}
\toprule
 &  &SQuADv2 &QA$^2$ & \data & BBC & Reddit & Yelp & Average \\ \midrule
\multirow{4}{*}{GPT-3.5} & ZS & 14.33 & 18.98 & 26.24 & \phantom{0}4.94 & 13.55 & 10.74 & 14.80 \\
 & ZS CoT & \phantom{0}7.27 & 11.18 & 14.30 & \phantom{0}3.60 & \phantom{0}7.52 & \phantom{0}3.97 & \phantom{0}7.97 \\
 & FS & \phantom{0}6.75 & \textit{20.46} & 33.78 & \phantom{0}0.00 & \phantom{0}8.66 & \phantom{0}9.47 & 13.19 \\
 & FS CoT & \textit{14.97} & 16.76 & 38.43 & \textit{\phantom{0}8.86} & \textit{21.90} & \textit{20.22} & \textit{20.19} \\\midrule
\multirow{4}{*}{GPT-4} & ZS & \textit{\textbf{57.92}} & \textit{\textbf{42.45}} & 44.52 & \textit{\textbf{64.72}} & 55.18 & \textit{\textbf{59.91}} & \textit{\textbf{54.12}} \\
 & ZS CoT & 44.51 & 33.66 & 31.64 & 51.80 & 32.54 & 42.49 & 39.44 \\
 & FS & 43.98 & 35.28 & 47.66 & 59.4 & 54.18 & 53.71 & 49.03 \\
 & FS CoT & 42.31 & 34.43 & \textit{48.84} & 60.27 & \textit{\textbf{59.93}} & 52.49 & 49.71 \\\midrule
\multirow{4}{*}{Llama2} & ZS & \phantom{0}8.26 & 16.42 & 23.30 & 17.22 & \phantom{0}8.66 & 15.94 & 14.97 \\
 & ZS CoT & \phantom{0}7.90 & \phantom{0}8.62 & 14.60 & \phantom{0}9.31 & \phantom{0}6.41 & 10.61 & \phantom{0}9.57 \\
 & FS & \textit{33.02} & \textit{35.71} & \textit{\textbf{51.55}} & \textit{28.46} & \textit{38.01} & \textit{38.38} & \textit{37.52} \\
 & FS CoT & \phantom{0}9.78 & 11.18 & 19.66 & 13.01 & 11.10 & 14.16 & 13.15 \\\midrule
\multirow{4}{*}{Mistral} & ZS & 10.62 & 10.12 & 15.42 & 16.88 & 10.71 & \phantom{0}8.19 & 11.99 \\
 & ZS CoT & \phantom{0}7.55 & \phantom{0}5.20 & 12.60 & \phantom{0}8.36 & \phantom{0}5.63 & \phantom{0}8.64 & \phantom{0}8.00 \\
 & FS & \textit{14.34} & 12.71 & 19.56 & 28.50 & \textit{36.13} & \textit{32.55} & \textit{23.97} \\
 & FS CoT & 12.64 & \textit{12.93} & \textit{19.64} & \textit{19.72} & 22.86 & 20.50 & 18.05 \\\midrule
\multirow{4}{*}{Zephyr} & ZS & \textit{31.48} & 27.49 & \textit{42.97} & \textit{38.23} & 22.54 & 20.85 & 30.59 \\
 & ZS CoT & 21.60 & 16.43 & 26.66 & 26.55 & 16.45 & 19.63 & 21.22 \\
 & FS & 27.89 & \textit{32.76} & 39.19 & 25.91 & 26.50 & 16.05 & 28.05 \\
 & FS CoT & 30.72 & 28.60 & 42.91 & 35.22 & \textit{42.45} & \textit{33.67} & \textit{35.59}    \\ \bottomrule
\end{tabular}
\caption{Cosine similarity using different prompting methods with both proprietary and open-source models on \benchmark. The best method for each base model is highlighted with an underscore, and the best method across all base models is bolded.}
\label{tab:cosine-similarity}
\end{table*}

\begin{table*}[t]
\centering
\small
\begin{tabular}{llccccccc}
\toprule
 &  &SQuADv2 &QA$^2$ & \data & BBC & Reddit & Yelp & Average \\ \midrule
\multirow{4}{*}{GPT-3.5} & ZS     & \textit{\textbf{\phantom{0}1.23}} & \textit{\textbf{\phantom{0}2.04}} & \textit{\textbf{\phantom{0}2.37}} & \textit{\textbf{\phantom{0}0.24}} & \phantom{0}0.95                   & \phantom{0}0.51                   & \textit{\textbf{\phantom{0}1.22}} \\
                         & ZS CoT & \phantom{0}4.88                   & 11.70                   & 11.58                  & \phantom{0}3.07                   & \phantom{0}3.35                   & \phantom{0}6.25                   & \phantom{0}6.81                   \\
                         & FS     & \phantom{0}1.79                   & \phantom{0}2.90                    & \phantom{0}4.24                   & -                      & \textit{\textbf{\phantom{0}0.94}} & \textit{\textbf{\phantom{0}0.37}} & \phantom{0}1.71                   \\
                         & FS CoT & \phantom{0}3.12                   & \phantom{0}2.77                   & \phantom{0}5.57                   & \phantom{0}1.22                   & \phantom{0}2.66                   & \phantom{0}2.20                    & \phantom{0}2.92                   \\\midrule
\multirow{4}{*}{GPT-4}   & ZS     & \textit{\phantom{0}4.09}          & \phantom{0}4.80                    & \textit{\phantom{0}5.63}          & \phantom{0}6.63                   & \textit{\phantom{0}5.69}          & \textit{\phantom{0}6.47}          & \textit{\phantom{0}5.55}          \\
                         & ZS CoT & 13.64                  & 16.77                  & 16.72                  & 15.92                  & 18.76                  & 22.27                  & 17.35                  \\
                         & FS     & \phantom{0}4.18                   & \textit{\phantom{0}4.53}          & \phantom{0}6.31                   & \textit{\phantom{0}6.46}          & \phantom{0}7.21                   & \phantom{0}7.78                   & \phantom{0}6.08                   \\
                         & FS CoT & \phantom{0}5.05                   & \phantom{0}5.86                   & \phantom{0}7.47                   & \phantom{0}8.41                   & \phantom{0}9.98                   & \phantom{0}8.29                   & \phantom{0}7.51                   \\\midrule
\multirow{4}{*}{Llama2}  & ZS     & \textit{\phantom{0}2.24}          & \phantom{0}4.87                   & \phantom{0}5.40                    & \textit{\phantom{0}1.47}          & \textit{\phantom{0}0.82}          & \textit{\phantom{0}1.02}          & \phantom{0}2.64                   \\
                         & ZS CoT & \phantom{0}6.61                   & 13.05                  & 15.10                   & \phantom{0}7.08                   & \phantom{0}4.70                    & \phantom{0}9.33                   & \phantom{0}9.31                   \\
                         & FS     & \phantom{0}2.53                   & \textit{\phantom{0}3.19}          & \textit{\phantom{0}2.66}          & \phantom{0}1.54                   & \phantom{0}2.13                   & \phantom{0}2.37                   & \textit{\phantom{0}2.40}           \\
                         & FS CoT & \phantom{0}9.02                   & 10.27                  & 18.04                  & \phantom{0}8.29                   & 12.09                  & \phantom{0}9.55                   & 11.21                  \\\midrule
\multirow{4}{*}{Mistral} & ZS     & \textit{\phantom{0}5.93}          & \textit{\phantom{0}5.20}           & \phantom{0}7.85                   & \textit{\phantom{0}3.29}          & \textit{\phantom{0}2.19}          & \textit{\phantom{0}2.18}          & \textit{\phantom{0}4.44}          \\
                         & ZS CoT & 11.20                   & \phantom{0}9.17                   & 20.88                  & 11.39                  & \phantom{0}9.21                   & \phantom{0}8.88                   & 11.79                  \\
                         & FS     & \phantom{0}6.05                   & \phantom{0}4.23                   & \textit{\phantom{0}6.00}             & \phantom{0}4.29                   & \phantom{0}4.69                   & \phantom{0}5.27                   & \phantom{0}5.09                   \\
                         & FS CoT & 16.05                  & 17.85                  & 25.98                  & 20.17                  & 24.45                  & 15.49                  & 20.00                     \\\midrule
\multirow{4}{*}{Zephyr}  & ZS     & 11.53                  & 12.28                  & 13.04                  & 10.93                  & \phantom{0}6.22                   & \textit{\phantom{0}3.41}          & \phantom{0}9.57                   \\
                         & ZS CoT & 40.03                  & 29.28                  & 47.48                  & 30.58                  & 15.26                  & 15.53                  & 29.69                  \\
                         & FS     & \textit{\phantom{0}8.32}          & \textit{\phantom{0}8.64}          & \textit{\phantom{0}9.70}           & \textit{\phantom{0}7.00}             & \textit{\phantom{0}6.06}          & \phantom{0}3.84                   & \textit{\phantom{0}7.26}          \\
                         & FS CoT & 14.95                  & 25.63                  & 22.88                  & 16.14                  & 13.13                  & \phantom{0}9.90  &17.11   \\ \bottomrule
\end{tabular}
\caption{Edit distance using different prompting methods with both proprietary and open-source models on \benchmark. The best method for each base model is highlighted with an underscore, and the best method across all base models is bolded.}
\label{tab:edit-distance}
\end{table*}
\section{More Annotation Details}
\begin{figure*}[t]
    \centering
    \includegraphics[width=\textwidth]{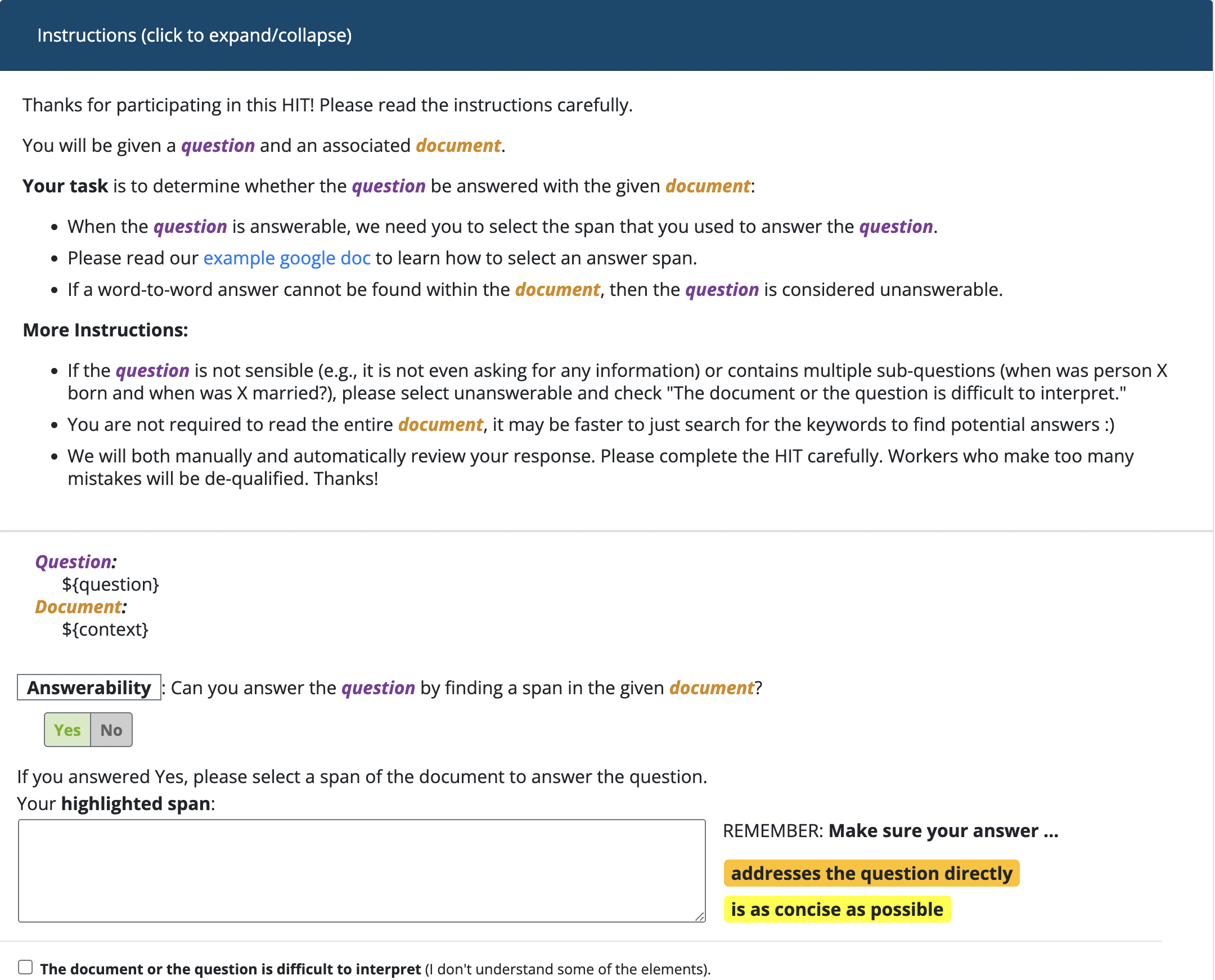}
    \caption{Annotation guideline for crowdworkers to annotate unanswerable questions.}
    \label{fig:annotation-guideline}
\end{figure*}

\paragraph{Annotation Guidelines} We present annotation guideline for annotating unanswerable questions in Figure~\ref{fig:annotation-guideline}. The crowdworkers are those who were identified to contribute high-quality annotations from our previous annotation tasks. For every completed HIT, we pay the crowdworker USD 0.5.

\paragraph{Annotation Agreement}
We report inter-annotator agreement rates between crowdworkers in Table~\ref{tab:human-agreement}. Specifically, we compute how often all three workers agree with each other. The agreement rates on the three datasets are close to each other.

\end{document}